\documentclass[runningheads]{llncs}
\usepackage{bibnames}
\usepackage{graphicx}
\usepackage{url}
\usepackage{cite}
\usepackage{csquotes}
\usepackage{float} 
\usepackage{enumitem}
\usepackage{hyperref}
\begin{document}
\title{Hate speech detection using static BERT embeddings \thanks{all authors contributed equally.}}
%
%
\author{Gaurav Rajput\and
Narinder Singh punn\and
Sanjay Kumar Sonbhadra \and
Sonali Agarwal
}
\authorrunning{Rajput et al.}
%
\institute{Indian Institute of Information Technology, Allahabad, Uttar Pradesh 211015, India
\email{gauravrajputoths@gmail.com, \{pse2017002, rsi2017502, sonali\}@iiita.ac.in}}
\maketitle              
\begin{abstract}
With increasing popularity of social media platforms hate speech is emerging as a major concern, where it expresses abusive speech that targets specific group characteristics, such as gender, religion or ethnicity to spread violence. Earlier people use to verbally deliver hate speeches but now with the expansion of technology, some people are deliberately using social media platforms to spread hate by posting, sharing, commenting, etc. Whether it is Christchurch mosque shootings or hate crimes against Asians in west, it has been observed that the convicts are very much influenced from hate text present online. Even though AI systems are in place to flag such text but one of the key challenges is to reduce the false positive rate (marking non hate as hate), so that these systems can detect hate speech without undermining the freedom of expression. In this paper, we use ETHOS hate speech detection dataset and analyze the performance of hate speech detection classifier by replacing or integrating the word embeddings (fastText (FT), GloVe (GV) or FT + GV) with static BERT embeddings (BE). With the extensive experimental trails it is observed that the neural network performed better with static BE compared to using FT, GV or FT + GV as word embeddings. In comparison to fine-tuned BERT, one metric that significantly improved is specificity.

\keywords{Hate speech detection \and BERT embeddings \and Word embeddings \and BERT.}
\end{abstract}
\section{Introduction}
With growing access to internet and many people joining the social media platforms, people tend to post online as per their desire and tag it as freedom of speech. It is one of the major problems on social media that tend to degrade the overall user's experience. Facebook defines hate speech as a direct attack against people on the basis of protected characteristics: race, ethnicity, national origin, disability, religious affiliation, caste, sexual orientation, sex, gender identity and serious disease \cite{facebook}, while for Twitter hateful conduct includes language that dehumanizes others on the basis of religion or caste \cite{twitter}. In March 2020, Twitter expanded the rule to include languages that dehumanizes on the basis of age, disability, or disease. Furthermore, the hateful conduct policy was expanded to also include race, ethnicity, or national origin \cite{twitter}. Following this context, hate speech can be defined as an abusive speech that targets specific group characteristics, such as gender, religion, or ethnicity.

Considering the massive amount of text what people post on social media, it is impossible to manually flag them as hate speech and remove them. Hence, it is required to have automated ways using artificial intelligence (AI) to flag and remove such content in real-time. While such automated AI systems are in place on social media platform, but one of the key challenges is the separation of hate speech from other instances of offensive language and other being higher false positive (marking non-hate as hate) rates of such system. Higher false positive rate means system will tag more non-hate content as hate content which can undermine the right to speak freely.

Hate speech can be detected using state-of-the-art machine learning classifiers such as logistic regression, SVM, decision trees, random forests, etc. However, deep neural networks (DNNs) such as convolutional neural networks (CNNs), long short-term memory networks (LSTMs) \cite{lstm_hochreiter1997long}, bidirectional long short-term memory networks (BiLSTMs) \cite{bilstm}, etc. have outperformed the former mentioned classifiers for hate speech detection \cite{mollas2020ethos}. Former classifiers do not require any word embedding \cite{word_embedding} to work with while the latter ones i.e DNNs requires word embeddings such as GloVe (GV) \cite{pennington2014glove}, fastText (FT) \cite{joulin2016fasttext}, Word2Vec \cite{mikolov2013distributed}, etc. Following this context, the present research work focuses on the scope of improvement of the existing state-of-the-art deep learning based classifiers by using static BERT \cite{devlin2018bert} embeddings (BE) with CNNs, BiLSTMs, LSTMs and gated recurrent unit (GRU) \cite{gru_cho2014learning}.

\subsection{BERT}
\begin{figure}[H]
  \includegraphics[width=\linewidth]{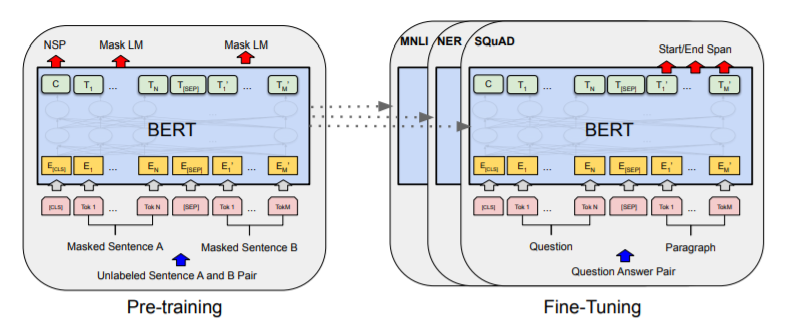}
  \caption{Pre-training and fine-tuning of BERT.}
  \label{fig:bert-training}
\end{figure}

Bidirectional encoder representations  from transformers (BERT) was developed by Devlin et al. \cite{devlin2018bert} in 2018. BERT is a transformer-based ML technique pre-trained from unlabeled data that is taken from Wikipedia (language: English) and BookCorpus. Transformer \cite{vaswani2017attention} model has two main parts: encoder and decoder. BERT is created by stacking the encoders. Two major strategies that BERT uses for training are masked language modelling (MLM) and next sentence prediction (NSP). The MLM strategy and fine-tuning of BERT is pictorially depicted in Fig \ref{fig:bert-training}.
In MLM technique 15\% of the words in a sentence are selected randomly and masked. Based on the context of the other words (which are not masked) the model tries to predict the masked word.

In NSP technique model is given pairs of sentences as input. The model learns to predict if the second sentence in a selected pair is the subsequent sentence in original document. During the training phase half of the inputs are a pair in which second sentence is subsequent sentence to the first in the original document while the rest half of the input pairs has a randomly selected sentence as second sentence.

\subsection{Attention in neural networks}
While processing a sentence in natural language for any NLP task, all words are not of equal importance, hence it is necessary to put more attention to the important words of the sentence. The importance of words depends on the context and is learned through training data. Bahdanau et al. \cite{bahdanau2014neural} proposed the attention mechanism for improving machine translation that uses seq-to-seq model. It is done by aligning the decoder with the relevant input sentences and implementing attention. Steps for applying attention mechanism are as follows:
\begin{enumerate}[label= \arabic*]
\item Produce the encoder hidden states.
\item Calculate alignment scores.
\item Soft-max the alignment scores.
\item Calculate the context vector.
\item Decode the output.
\item At each time step, based on the current state of decoder and input received by decoder, an element of decoder's output sequence is generated by decoder. Besides that decoder also updates its own state for next time step. Steps 2-5 repeats itself for each time step of the decoder until the output length exceeds a specified maximum length or end of sentence token is produced.
\end{enumerate}

The rest of the paper is organised as follows: literature review in Section 2 which focuses on the related work and recent developments in this field, Section 3 describes the proposed methodology. Section 4 covers the exhaustive experimental trials followed by improved results in section 5, and lastly in Section 6 the concluding remarks are presented.

\section{Related work}
The advancements in deep learning technology have widen spectrum of its application tasks involving classification, segmentation, object detection, etc., across various domains such as healthcare, image processing, natural language processing, etc. \cite{punn2021automated, punn2020inception, batra2021bert, punn2020multi, zhang2019empirical}. With hate speech detection being one of the major problems in the evergrowing social media platforms, it has drawn keen interest of the research community to develop AI assisted applications. Following this, Badjatiya et al. \cite{badjatiya2017deep} proposed a deep learning approach to perform hate speech detection in tweets. The approach was validated on the dataset \cite{waseem2016hateful} that consists of 16,000 tweets, of which 1972 are marked as racist, 3383 as sexist and the remaining ones as neither. The authors utilized convolutional neural networks, long short-term memory networks and FastText. The word embeddings are initialized with either random embeddings or GV \cite{pennington2014glove} embeddings. The authors achieved promising results with “LSTM + Random Embedding + GBDT” model. In this model, the tweet embeddings were initialized to random vectors, LSTM was trained using back-propagation, and then learned embeddings were used to train a GBDT classifier.
 
Rizos et al. \cite{rizos2019augment} experimented by using short-text data augmentation technique in deep learning for hate speech classification. For short-text data augmentation they used  substitution based augmentation (ThreshAug), word position augmentation (PosAug) and neural generative augmentation (GenAug). For performing experiments they used the dataset \cite{davidson2017automated} which consists of around 24k samples, of which 5.77\% samples are marked as hate, 77.43\% samples are marked as offensive and 16.80\% samples as neither. The authors experimented with multiple DNNs such as CNN, LSTM and GRU. In addition, fastText, GloVe and Word2Vec were used as word embeddings. They achieved their best results by using GloVe + CNN + LSTM + BestAug, where BestAug is combination of PosAug and ThreshAug. Faris et al. \cite{faris2020hate} proposed a deep learning approach to detect hate speech in Arabic language context. They created their dataset by scraping tweets from twitter using an application programming interface (API) \cite{api_wikipedia_2021} and performed standard dataset cleaning methods. The obtained dataset have 3696 samples of which 843 samples are labelled as hate and 791 samples as normal while rest of the samples were labelled as neutral. Word2Vec and AraVec \cite{soliman2017aravec} were used for feature representation and embedding dimension was kept to 100. The authors achieved promising results using combination of CNN and LSTM with AraVec.

Ranasinghe et al. \cite{ranasinghe2019brums} in hate
speech and offensive content identification in Indo-European languages
(HASOC) shared task 2019 experimented with multiple DNNs such as pooled GRU, stacked LSTM + attention, LSTM + GRU + attention, GRU + capsule using fastText as word embedding on the dataset having posts written in 3 languages: German, English and code-mixed Hindi. Furthermore, they also experimented with fine-tuned BERT \cite{devlin2018bert} which outperformed every above mentioned DNN for all 3 languages. In another work, Mollas et al. \cite{mollas2020ethos} proposed ETHOS dataset to develop AI based hate speech detection framework that have used FT \cite{joulin2016fasttext}, GV \cite{pennington2014glove} and FT + GV as word embeddings with CNNs, BiLSTMs and LSTMs. In contrast to other datasets which are based on tweets scraped from Twitter, this new dataset is based on YouTube and Reddit comments. A binary version of ETHOS dataset has 433 sentences containing hate text and 565 sentences containing non-hate text. Besides, transfer learning was used to fine-tune BERT model on the proposed dataset that outperformed the above mentioned deep neural networks. The results of the aforementioned experiments are shown in Table \ref{table1}, where bold values represent the highest value of metrics among all models \cite{mollas2020ethos}.

\begin{table}[H]
\centering
\caption{Performance of BERT (fine-tuned on binary ETHOS dataset) with various neural networks using FT, GV or FT + GV as word embedding \cite{mollas2020ethos}}\label{table1}
\begin{tabular}{|c|c|c|c|c|c|}
\hline
 Model &  F1 Score & Accuracy & Precision  & Recall & Specificity\\
\hline
CNN + Attention + FT + GV &	74.41 &	75.15 &	74.92 & 74.35 &	{\bfseries 80.35}\\
\hline
CNN + LSTM + GV	& 72.13	& 72.94	& 73.47	& 72.4	& 76.65\\
\hline
LSTM + FT + GV	& 72.85	& 73.43	& 73.37	& 72.97	& 76.44\\
\hline
BiLSTM + FT + GV	& 76.85	& {\bfseries 77.45} &	77.99	& 77.10	& 79.66\\
\hline
BiLSTM + Attention + FT	& 76.80	& 77.34	& 77.76	& 77.00	& 79.63\\
\hline
BERT	& {\bfseries 78.83} &	76.64	& {\bfseries 79.17} & 	{\bfseries 78.43} &	74.31\\
\hline
\end{tabular}
\end{table}

Ever since the researchers started using BERT \cite{devlin2018bert} for natural language processing tasks, it has been observed that a fine-tuned BERT usually outperforms other state-of-the-art deep neural networks in same natural language processing task. The same has been observed in the results of experiment carried out by Mollas et al \cite{mollas2020ethos}. Motivated from this, the experiments carried out in this paper aims to analyse the performance of fine-tuned BERT with other deep learning models.

\section{Proposed methodology}
\begin{figure}[H]
  \includegraphics[width=\linewidth]{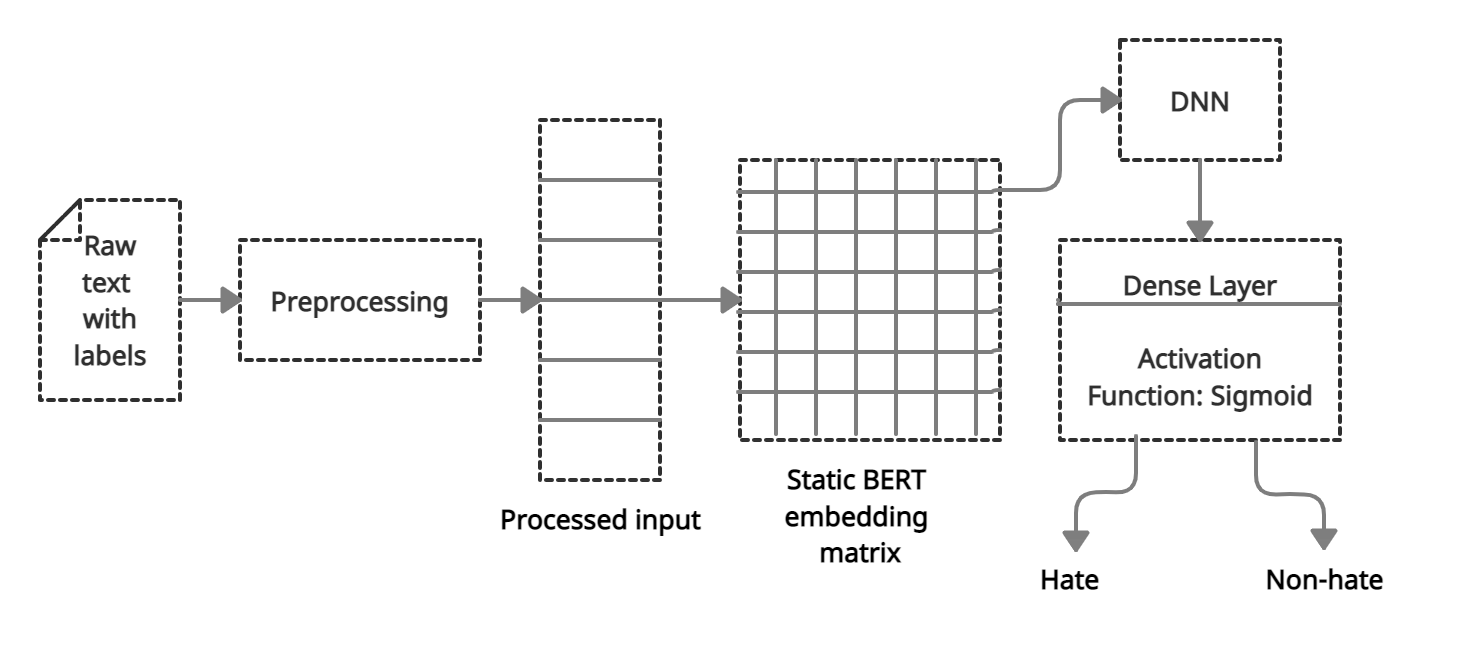}
  \caption{Block diagram of proposed model.}
  \label{fig:block_diagram}
\end{figure}

Following the state-of-the-art deep learning classification models, in the proposed approach the impact of BERT based embeddings is analyzed. The hate speech detection framework is designed by combining DNNs (CNN, LSTM, BiLSTM and GRU) with static BERT embeddings to better extract the contextual information. Initially, the static BERT embedding matrix is generated from large corpus of dataset, representing embedding for each word and later, this matrix is processed using DNN classifiers to identify the presence of hate. The schematic representation of the proposed model is shown in Fig \ref{fig:block_diagram}.

\subsection{Static BERT embedding matrix}
The embedding matrix contains the word embeddings for each word in dataset. Each row in the embedding matrix contains word embedding for a unique word and they are passed to the DNNs (by converting natural language sentences to vectors) that accepts input in fixed dimensions, therefore the word embeddings have to be static. Since BERT \cite{devlin2018bert} gives contexualised embedding of each word according to the usage of the word in sentence, thereby same word will have different embeddings depending on the usage context unlike in other static word embeddings where each word has unique static embedding irrespective of the context in which it is used. 

Initially, the raw BERT embeddings are generated using bert-embedding library \cite{pypi} to provide contextualized word embedding. An embedding dictionary (key-value pair) is developed where key is the unique word and value is an array containing contextualized embeddings of that unique word. Since same word can be used in different context in different sentences, hence it will have different word embeddings depending on the context. Every contextualized embedding for a word are stored in the dictionary \cite{python_dict} by pushing the embedding into the vector corresponding to the unique word. Furthermore, the static BE of a word is obtained by taking mean of the vector containing the contexualized  BERT embeddings of that word. For example, a word \lq\textit{W}\rq occurs 4 times in the dataset, then there will be 4 contexualized embeddings of \lq\textit{W}\rq, let it be {\itshape $E_1, E_2, E_3, E_4$}. These 4 embeddings each of dimension (768,) are stored in the array corresponding to the key \lq\textit{W}\rq in the dictionary. Later, mean of {\itshape $E_1, E_2, E_3, E_4$} is computed that represents the static BERT \cite{devlin2018bert} embedding of \lq\textit{W}\rq. For words which are not in vocabulary, BERT \cite{devlin2018bert} splits them into subwords and generate their embeddings, then take the average of embeddings of subwords to generate the embedding of the word which was not in vocabulary. Finally, by using keras Tokenizer \cite{keras-team} and static BERT embeddings we create the embedding matrix.

\section{Experiments}

\subsection{Choice of dataset}
Even though there are multiple datasets that are publicly available for hate speech detection but we chose to use binary version of ETHOS dataset \cite{mollas2020ethos} because it is the most recent dataset and the two classes (hate and non-hate) present in it are almost balanced as compared to other datasets. For example, Davidson dataset having around 24k samples (Hate speech: 5.77\%, Offensive: 77.43\% and 16.80\% as Neither) \cite{davidson2017automated} is highly imbalanced. ETHOS dataset address all such issues of available datasets.

The Shannon entropy can be used as a measure of balance for datasets. On a dataset of $n$ instances, if we have $k$ classes of size $c_i$ we can compute entropy as follows:
\begin{equation}
H\;=\;-\sum_{i\;=\;1}^k\frac{c_i}n\log\frac{c_i}n
\end{equation}
It is equal to zero if there is a single class. In other words, it tends to 0 when the dataset is very unbalanced and $log (k)$  when all the classes are balanced and of the same size $n/k$. Therefore, we use the following measure of balance (shown in Eq. \ref{eq1}) for a dataset \cite{cross_validated_2016}:
\begin{equation}
Balance =\frac H{\log\;k}=\frac{-\sum_{i=1}^k\frac{c_i}n\log{\frac{c_i}n}}{\log\;k}
\label{eq1}
\end{equation}
 
 Binary version of ETHOS dataset has 433 samples containing hate text and 565 samples containing non-hate text. For binary version of ETHOS dataset, $Balance = 0.986$ which is nearly equal to 1, indicating balance between classes.

\subsection{Neural network architectures and testing environment}
The proposed approach is trained and validated on the binary version of ETHOS dataset. For the purpose of comparison, the neural network architectures are kept exactly same as described by the Mollas et al. \cite{mollas2020ethos}. From the number of units in a neural network to the arrangement of layers in a neural network everything is kept same so as to create the same training and testing environment but change the word embeddings to static BERT embeddings.

To establish robust results, stratified k-fold validation technique with value of $k = 10$ is utilized. Furthermore, the training phase is assisted with callbacks such as early stopping (stop the training if performance doesn't improve) to avoid the overfitting problem and model-checkpointing (saving the best model). Finally, the trained model is evaluated using standard classification performance metrics i.e. accuracy, precision, recall (sensitivity), F1-score and specificity.

\begin{equation}
Accuracy = \frac{TP\;+\;TN}{TP\;+\;TN\;+\;FP\;+\;FN}
\end{equation}

\begin{equation}
Precision = \frac{TP\;}{TP\;+FP}    
\end{equation}

\begin{equation}
Recall = \frac{TP\;}{TP\;+FN}    
\end{equation}

\begin{equation}
F1-score = 2\;\times\frac{Precision\;\times\;Recall}{Precision\;+\;Recall}    
\end{equation}

\begin{equation}
Specificity = \frac{TN\;}{TN\;+FP}   
\end{equation}
Where,
\textit{TP}: True Positive, \textit{TN}: True Negative,  \textit{FP}: False Positive, \textit{FN}: False Negative

\section{Results and discussion}

Table \ref{table1} represents the results of experiment carried out by Mollas et al. \cite{mollas2020ethos} on binary version of ETHOS dataset, hence the DNNs uses only FT, GV  or FT + GV as word embeddings. It is evident from the Table \ref{table1} that BERT (fine-tuned on binary ETHOS dataset) outperformed other models in all metrics except accuracy and specificity, its specificity stands at 74.31\% which indicates high false positive hate speech classification.

\begin{table}[H]
\centering
\caption{Comparative analysis of the performance of various DNNs with and without static BERT embeddings (BE).}\label{table2}
\begin{tabular}{|c|c|c|c|c|c|}
\hline
 Model &  F1-Score & Accuracy & Precision & Recall & Specificity\\
\hline
\hline
CNN + Attention + FT + GV	& 74.41	& 75.15	& 74.92	& 74.35	& 80.35\\
\hline
\bfseries{CNN + Attention + static BE}	& 77.52	& 77.96	& 77.89	& 77.69	& 79.62\\
\hline
\hline
CNN + LSTM + GV	& 72.13	& 72.94	& 73.47	& 72.4	& 76.65\\
\hline
\bfseries{CNN + LSTM + static BE}	& 76.04	& 76.66	& 77.20	& 76.18	& 79.43\\
\hline
\hline
LSTM + FT + GV	& 72.85	& 73.43	& 73.37& 72.97	& 76.44\\
\hline
\bfseries{LSTM + static BE}	& 79.08	& 79.36	& 79.38 & 79.37	& 79.49\\
\hline
\hline
BiLSTM + FT + GV	& 76.85	& 77.45	& 77.99	& 77.10	& 79.66\\
\hline
\bfseries{BiLSTM + static BE}	& {\bfseries 79.71}	& {\bfseries 80.15}	& {\bfseries 80.37}	& {\bfseries 79.76} & {\bfseries 83.03}\\
\hline
\hline
BiLSTM + Attention + FT	& 76.80	& 77.34	& 77.76& 77.00	& 79.63\\
\hline
\bfseries{BiLSTM + Attention+static BE}	& 78.52	& 79.16	& 79.67 & 78.58	& 83.00\\
\hline
\hline
\bfseries{GRU + static BE}	& 77.91	& 78.36	& 78.59	& 78.18	& 79.47\\
\hline
BERT	& 78.83	& 76.64	& 79.17 & 78.43	& 74.31\\
\hline
\end{tabular}
Bold model names represent static BERT embedding variants of the models
\newline
Bold values represent the highest value of any metric among all models 
\end{table}

The Table \ref{table2} presents the obtained results on various DNNs, where bold model names represent BERT variant of a DNN model and bold quantities represent the highest values. It is observed that a deep neural network with static BERT embeddings outperforms the same deep neural network which is using word embedding as fastText, GloVe or fastText + GloVe in all metrics. For DNNs like CNN using attention, LSTM, CNN + LSTM, BiLSTM and BiLSTM using attention, the average (avg) increase in F1-score is 3.56\%, accuracy is 3.39\%, precision is 3.40\%, recall is 3.55\% and sensitivity is 2.37\%. Hence, it is evident that static BERT embeddings tend to provide better feature representation as compared to fastText, GloVe or fastText + GloVe. 

Furthermore, BiLSTM using static BERT embeddings (BiLSTM + static BE) performs better in all metrics as compared to other DNNs under consideration. In the results of experiments done by Mollas et al. \cite{mollas2020ethos}, fine-tuned BERT outperformed other models in every metric with specificity as 74.31\%, which increases to 83.03\% using static BERT embeddings (BiLSTM + static BE), thereby attaining a significant increase of 8.72\%.

\section{Conclusion}
In this article, the impact of performance in deep learning based hate speech detection using static BERT embeddings is analysed. With exhaustive experimental trials on various deep neural networks it is observed that using neural networks with static BERT embeddings can significantly increase the performance of the hate speech detection models especially in terms of specificity, indicating that model is excellent at correctly identifying non hate speeches. Therefore, it flags lesser non hate speech as hate speech, thereby protecting the freedom of speech. With such promising improvements in the results, the same concept of integrating static BERT embedddings with state-of-the-art models can further be extended to other natural language processing based applications.

\bibliographystyle{splncs04}
\bibliography{reference}

\end{document}